%% file: main.tex
\documentclass[letterpaper, 10 pt, conference]{ieeeconf}  

\input{packages}

\input{commands}

\IEEEoverridecommandlockouts                              

\title{\LARGE \bf A Slices Perspective for Incremental Nonparametric Inference in High Dimensional State Spaces
}

\author{Moshe Shienman$^{1}$, Ohad Levy-Or$^{1}$, Michael Kaess$^{2}$ and Vadim Indelman$^{3}$
\thanks{$^{1}$Moshe Shienman and Ohad Levy-Or are with the Technion Autonomous Systems Program (TASP), Technion - Israel Institute of Technology, Haifa 32000,
	Israel, {\tt\small \{smoshe,ohadlor\}@campus.technion.ac.il}}
\thanks{$^{2}$Michael Kaess is with the Robotics Institute, Carnegie Mellon University, Pittsburgh, PA 15213, USA. {\tt\small kaess@cmu.edu}}
\thanks{$^{3}$Vadim Indelman is with the Department of Aerospace Engineering, Technion - Israel Institute of Technology, Haifa 32000, Israel. {\tt\small vadim.indelman@technion.ac.il}}
\thanks{*This work was  partially supported by US NSF/US-Israel BSF.}
}

\begin{document}

\maketitle
\thispagestyle{empty}
\pagestyle{empty}

\input{00-abstract}

\input{01-introduction}

\input{02-relatedwork}

\input{03-formulation}

\input{04-methodology}

\input{05-results}

\input{06-conclusions}

\addtolength{\textheight}{-12cm}   

\input{07-appendix}


\bibliographystyle{plain}
\bibliography{../../../references/refs}

\end{document}

%% file: packages.tex
\usepackage{cite}
\usepackage{amsmath,amssymb,amsfonts}
\usepackage{algorithmic}
\usepackage{graphicx}
\usepackage{textcomp}
\usepackage{xcolor}
\usepackage{caption}
\usepackage{subcaption}
\usepackage{letltxmacro}
\makeatletter
\def\maketag@@@#1{\hbox{\m@th\normalfont\normalsize#1}}
\makeatother

\usepackage[linesnumbered,ruled,vlined]{algorithm2e}



%% file: commands.tex
\newtheorem{lemma}{Lemma}

\newcommand{\prob}[1]{\ensuremath{\mathbb{P}({#1})}}

\newcommand{\slices}{\ensuremath{\textit{slices}}\xspace}
\newcommand{\slice}{\ensuremath{\textit{slice}}\xspace}
\newcommand{\forward}{\ensuremath{\textit{forward}\;}}
\newcommand{\backward}{\ensuremath{\textit{backward}\;}}
\newcommand{\fb}{\ensuremath{\textit{forward-backward}\;}}
\newcommand{\mixture}{\ensuremath{\textit{mixture}\;}}

\SetCommentSty{mycommfont}

%% file: 00-abstract.tex
\begin{abstract}
We introduce an innovative method for incremental nonparametric probabilistic inference in high-dimensional state spaces. Our approach leverages \slices from high-dimensional surfaces to efficiently approximate posterior distributions of any shape. Unlike many existing graph-based methods, our \slices perspective eliminates the need for additional intermediate reconstructions, maintaining a more accurate representation of posterior distributions. Additionally, we propose a novel heuristic to balance between accuracy and efficiency, enabling real-time operation in nonparametric scenarios. In empirical evaluations on synthetic and real-world datasets, our \slices approach consistently outperforms other state-of-the-art methods. It demonstrates superior accuracy and achieves a significant reduction in computational complexity, often by an order of magnitude.
\end{abstract}

%% file: 01-introduction.tex
\section{INTRODUCTION}
	
	
	
	
Modern mobile robots and autonomous systems are expected to reliably and efficiently perform various tasks including navigation, exploration, planning, manipulation and tracking. Furthermore, they must operate under different sources of uncertainty such as noisy measurements, imprecise actions, and dynamic environments in which some events are unpredictable. When uncertainty plays a significant role and probabilistic reasoning is essential for making safe and informed decisions, access to the full posterior distribution, for both the robot and the environment state, is crucial. As such, probabilistic inference algorithms, providing posterior distribution estimates, are commonly used to address these challenges. However, handling high-dimensional state spaces and operating in real-time is far from trivial.
 
Substantial research efforts have been made in recent decades to develop probabilistic inference algorithms that are robust, accurate and capable of real time performance. These efforts often rely on various assumptions to achieve their objectives. Perhaps the most frequently employed assumption is that the actual posterior distribution can be approximated using a parametric Gaussian model. However, in real-world problems, the posterior distribution is often non-Gaussian, having multiple modes or a nonparametric structure. Due to the complex, non-Gaussian nature of such posterior distributions, obtaining closed-form analytical solutions is challenging and frequently impractical.

\begin{figure} [t]
	\begin{subfigure}[b]{0.10\textwidth}
		\centering
		\includegraphics[scale=0.35]{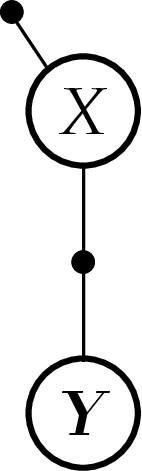}
		\caption{}
	\end{subfigure}
	\begin{subfigure}[b]{0.42\textwidth}
		\centering
		\includegraphics[scale=0.25]{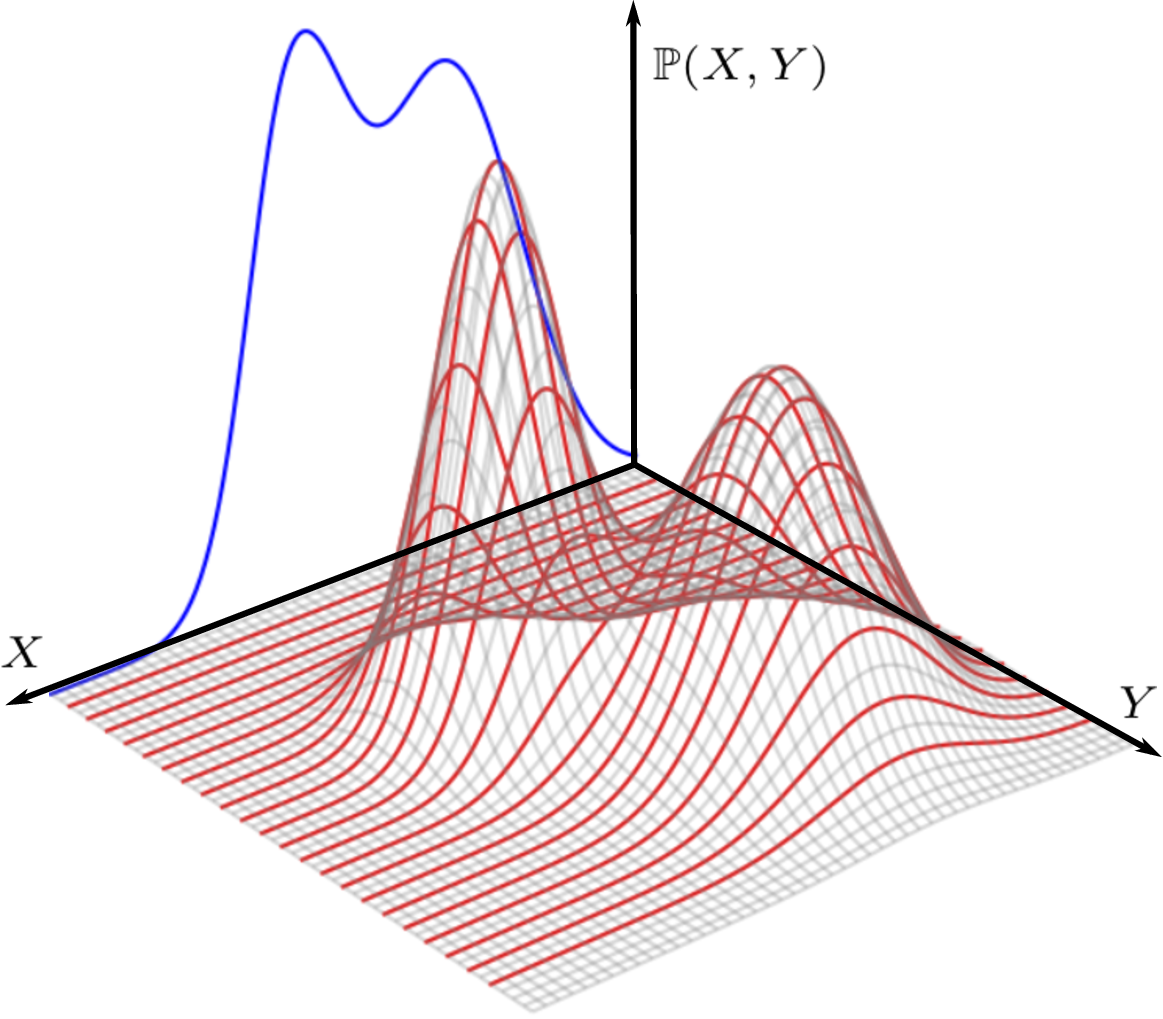}
		\caption{}
	\end{subfigure}
	\caption{\scriptsize (a) The factor graph of a small probabilistic inference problem containing two variable nodes $X,Y$ and two factor nodes $f(X), f(X,Y)$ indicated as small solid black circles; (b) A high-dimensional surface representing the joint distribution $\prob{X,Y}$. Each \textcolor{red}{red} \slice, at a specific realization $Y=y$, represents a conditional distribution $\prob{X | Y=y}$. The marginal distribution $\prob{X}$, shown in \textcolor{blue}{blue}, is calculated by integrating over all conditional \slices.}
	\label{fig: surface with slices}
\end{figure}

To approximate and model such non-Gaussian posterior distributions, modern nonparametric methods use different variants of the \fb algorithm, exploiting graphical models such as factor graphs \cite{Kschischang01it} and Bayes trees \cite{Kaess12ijrr}. In these methods, samples are generated at each step to reconstruct intermediate distributions through techniques such as Kernel Density Estimators (KDE) and various learning procedures.

Our key observation is that these distributions can be directly reconstructed without the need for any additional learning techniques or KDE. This is accomplished by accessing \slices from high-dimensional surfaces that represent partial joint distributions (Fig. \ref{fig: surface with slices}). By circumventing these processes, our approach proves to be computationally more efficient, requires significantly less samples and produces more accurate results, as supported by our experimental findings. 

Our main contributions in this paper are as follows: 
\begin{enumerate}
	\item We introduce a nonparametric inference approach which leverages \slices from high-dimensional surfaces to approximate joint and marginal posterior distributions without any further intermediate reconstructions.
	
	\item Unlike previous methods, ours does not require any iterative procedures nor breaking the underlying graphical model to generate samples in each step, even in cases where unary factors are not available.
	
	\item We show how to utilize our \slices perspective for nonparametric incremental inference and propose a novel early stopping heuristic criteria to further speed up calculations.
	
	\item Our approach requires less samples and consistently outperforms state-of-the-art nonparametric inference algorithms in terms of accuracy and computational complexity. This superiority is evident across evaluations conducted on both synthetic and real-world datasets, with improvements in time complexity reaching up to an order of magnitude. 
\end{enumerate}
Our proposed inference approach is general and can be effectively employed in various estimation problems such as tracking, sensor fusion, Bundle Adjustment (BA), Structure from Motion (SfM) and Simultaneous Localization and Mapping (SLAM). In this particular study, we focus on demonstrating our approach within the framework of SLAM.


%% file: 02-relatedwork.tex
\section{RELATED WORK}
\begin{figure*} [t]
	\centering
	\begin{subfigure}[b]{0.45\textwidth}
		\centering
		\includegraphics[scale=0.40]{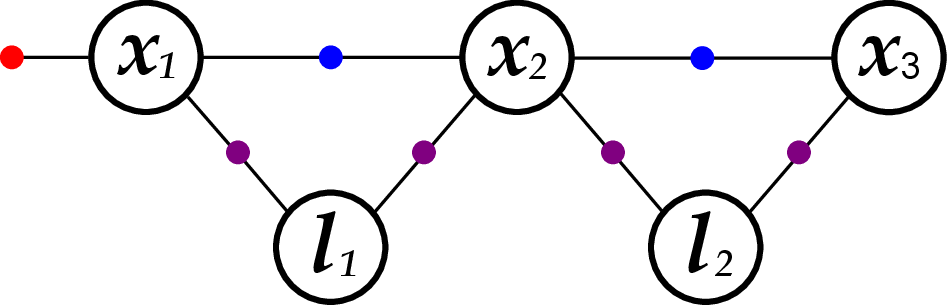}
		\caption{Factor graph}
		\label{fig: factor graph}
	\end{subfigure}
	\begin{subfigure}[b]{0.45\textwidth}
		\centering
		\includegraphics[scale=0.40]{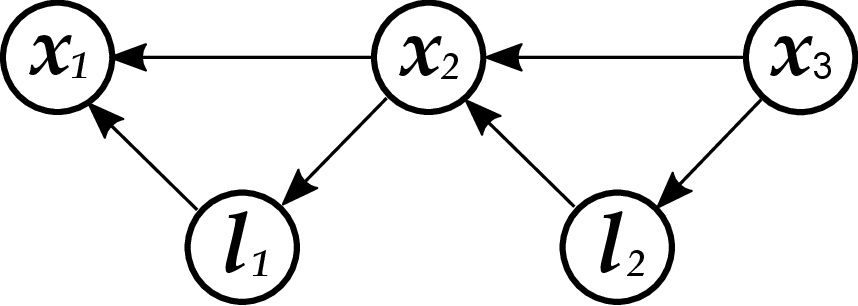}
		\caption{Bayes net}
		\label{fig: bayes net}
	\end{subfigure}
	\caption{(a) A factor graph formulation of a SLAM problem with five variable nodes and seven factor nodes. Factor nodes represent probabilistic information over random variables. In this example, factor nodes include: a prior \textcolor{red}{$f(x_1)$}, odometry measurements \textcolor{blue}{$f(x_1,x_2), f(x_2,x_3)$} and landmark measurements \textcolor{violet}{$f(x_1,l_1), f(x_2,l_1), f(x_2,l_2), f(x_3,l_2)$}. The factor graph represents a factorization of the joint distribution as a product of all factors;
		(b) The corresponding Bayes net after performing variable elimination on the factor graph using the elimination order $\mathcal{O} = \{\theta_1=x_1, \theta_2=l_1, \theta_3=x_2, \theta_4=l_2, \theta_5=x_3\}$. The joint distribution is expressed as the product of conditionals produced in each step as $\prob{x_1,x_2,x_3,l_1,l_2} = \prob{x_1|x_2,l_1} \cdot \prob{l_1|x_2} \cdot \prob{x_2|x_3,l_2} \cdot \prob{l_2|x_3} \cdot \prob{x_3}$.}	
	\label{fig: factor graph with bayes net}
\end{figure*}

Research on inference algorithms, specifically used for SLAM, has predominantly focused on parametric solutions, often relying on approximating posterior distributions with parametric Gaussian models. 
One such notable work is iSAM2 \cite{Kaess12ijrr}, which introduced the Bayes tree graphical model. iSAM2 recovers posterior distributions using incremental upward and downward passes over the Bayes tree, in similar manner to the \fb algorithm. iSAM2 also incorporates an early stopping heuristic in the downward pass, aimed at enhancing computational efficiency. 

Parametric approaches have more recently also been adapted to improve robustness, addressing challenges like unresolved data associations, multi-modal factors, and outliers \cite{Hsiao19icra, Huang13icassp, Indelman16csm, Indelman14icra, Olson13ijrr, Sunderhauf12iros}. Nevertheless, these methods cannot handle non-Gaussian posterior distributions.

Nonparametric inference methods utilize sampling techniques and can theoretically be used to approximate any posterior distribution. Markov Chain Monte Carlo (MCMC) algorithms \cite{Hastings70} and particle filters \cite{Gordon93} were among the initial paradigms explored in the development of such methods. However, these approaches often become computationally impractical in high-dimensional settings, such as those encountered in SLAM. FastSLAM \cite{Montemerlo02aaai} introduced a method that combines the strengths of particle filters and parametric techniques to address high-dimensional settings effectively. Nevertheless, it has limitations. FastSLAM uses extended Kalman filters to track landmarks and cannot represent general non-Gaussian distributions. Additionally, similar to many particle filters, FastSLAM can encounter the issue of particle depletion, which can result in degenerate estimates. 

Several recently proposed state-of-the-art methods leverage the conditional independence structure within factor graphs \cite{Kschischang01it} to handle nonparametric inference in SLAM. Among them, mm-iSAM \cite{Fourie16iros} and NF-iSAM \cite{Huang23tro} take advantage of the Bayes tree \cite{Kaess10wafr} to reduce computational complexity. In contrast, NSFG \cite{Huang22ral}, which employs nested sampling \cite{Skilling06}, sacrifices computational efficiency for higher accuracy. We concentrate on these three approaches because they align conceptually with our own method, given that we also utilize the factor graph model.

mm-iSAM approximates non-Gaussian posterior distributions through the use of samples and KDEs. More specifically, in each step, the mm-iSAM approximation process employs an iterative nested Gibbs sampling approach \cite{Ihler03nips} to generate samples from partial posteriors associated with each clique in the Bayes tree. These generated samples are then employed to construct new approximations of posterior distributions using KDEs. This distinguishes it conceptually from our approach, as we circumvent the need for an additional intermediate approximation with KDEs by directly accessing high-dimensional surfaces through sampled \slices. 

NF-iSAM exploits the expressive power of neural networks, and trains normalizing flows \cite{Rezende15icml} to model conditionals that factorize non-Gaussian posterior distributions during the upward pass. Once all cliques have learned their conditional samplers, NF-iSAM conducts a downward pass, drawing samples from posterior distributions in a root-to-leaf manner. However, this learning approach is computationally intensive  and requires many samples for convergence.

NSFG leverages nested sampling techniques to sample the posterior distribution directly across iterations until convergence. In this iterative process, the factor graph is initially divided into two components: an acyclic graph, referred to as the prior factor set, and a likelihood factor set. NSFG subsequently employs ancestral sampling to produce samples from the prior factor set, while likelihood evaluations are provided based on the likelihood factor set. Although NSFG delivers more precise estimates of non-Gaussian posterior distributions when compared to mm-iSAM and NF-iSAM, its iterative nature imposes significant computational demands. Consequently, it struggles to address high-dimensional, large-scale problems and cannot be employed in real-time applications. 

Our proposed approach, which utilizes the \slices perspective, demonstrates superior computational efficiency by eliminating the need for iterative procedures such as Gibbs sampling \cite{Fourie16iros}, training neural networks \cite{Huang23tro}, or NSFG \cite{Huang22ral}, which inherently involves iterations. Moreover, our approach exhibits accuracy on par with NSFG, which is considered as an offline algorithm.

%% file: 03-formulation.tex
\section{BACKGROUND AND NOTATIONS} \label{sec: background}
Consider a SLAM framework in which an autonomous agent is operating in a partially known environment. At each time step $k$ the agent takes an action $u_k$ and acquires an observation $Z_k \triangleq \{z_k^1,..,z_k^n \}$ based on $n$ measurements. The motion and observation models are given by
\begin{equation} \label{eq: models}
	x_{k+1} = f \left( x_k, u_k, w_k \right) \;,\; z_k = h \left( x_k , x^l, v_k \right) ,
\end{equation}
where $x^l$ denotes a landmark pose and $w_k, v_k$ are noise terms, sampled from known motion and measurement distributions, respectively. In this work specifically, while the motion and measurement distributions are known, they can have any arbitrary shape, i.e. not necessarily Gaussian. 

We denote the set of all state variables, including all agent poses and observed landmarks, by $\Theta$. 
Given all actions and measurements, the joint probability density function (the belief) is given by 
\begin{equation} \label{eq: belief}
	b_k[\Theta]\triangleq\prob{\Theta | u_{0:k-1}, Z_{0:k}} = \prob{\Theta | D},  
\end{equation}
where $D \triangleq \{b_0, u_{0:k-1}, Z_{0:k}\}$ represents all available data at time instant $k$. To reduce clutter, we omit time notations from hereon, and refer to the above joint density as $\prob{\Theta | D}$.
Our goal is to compute the joint posterior distribution \eqref{eq: belief} and marginal posterior distributions $\prob{\theta | D} \;,\; \forall \theta \in \Theta$. 

We use a factor graph model \cite{Kschischang01it} to represent the joint posterior distribution (Fig. \ref{fig: factor graph}). A factor graph $G = \left( \mathcal{F}, \Theta, \mathcal{E} \right)$ 
is a bipartite graph with two types of nodes: factor nodes $f \in \mathcal{F}$ and variable nodes $\theta \in \Theta $. The variable nodes represent the random variables in the estimation problem, whereas the factor nodes represent
probabilistic information on those variables. The edges in $\mathcal{E}$ encode connectivity based on the variables associated with each factor, with each edge $e (f, \theta) \in \mathcal{E}$ linking a factor node to a variable node.
This probabilistic graphical model represents a factorization of the posterior distribution in terms of process and measurement models
\begin{equation} \label{eq: posterior distribution}
	\prob{\Theta | D} \propto \prod_{i} f_i \left( \Theta_i \right),
\end{equation}
where $\Theta_i = \{ \theta \; | \; \exists e(f_i, \theta) \in \mathcal{E} \}$ is the set of all variables adjacent to factor $f_i$. 
In this work we assume for simplicity that only unary and pairwise factors exist in $G$, i.e. each $f \in \mathcal{F}$ can be connected with up to two variable nodes. 

A common approach for computing the posterior distribution \eqref{eq: belief} and the posterior marginals $\prob{\theta | D} \;,\; \forall \theta \in \Theta$ is the \fb algorithm which operates in two passes given a specific variable ordering $\mathcal{O} \triangleq \text{ord} (\Theta)$.

During the \forward pass, the factor graph is gradually transformed into a Bayes net \cite{Pearl88book} through a bipartite elimination game \cite{Heggernes96siam}. In each step, a single variable is eliminated from the factor graph following the elimination order $\mathcal{O}$, starting with a variable linked to a prior factor. Specifically, each step begins with a factor graph $G_{j-1} = \left( \mathcal{F}_{j-1}, \Theta_{j-1}, \mathcal{E}_{j-1} \right)$, where $G_0 = G$ by convention. A variable node $\theta_j$, representing the $j$th variable in $\mathcal{O}$, and all factors $\mathcal{F}_{j-1}(\theta_j) \triangleq \{f \; | \; \exists e(f, \theta_j) \in \mathcal{E}_{j-1} \}$ are first removed from $G_{j-1}$ along with the corresponding edges, i.e.  
$\mathcal{E}_{j-1}(\theta_j) \triangleq \{e(f, \theta) \; | \; f \in \mathcal{F}_{j-1}(\theta_j) , \theta \in \Theta_{j-1}\}$. All variables involved in $\mathcal{E}_{j-1}(\theta_j)$, except for $\theta_j$, define a separator $S_j$ representing the Markov blanket of variable node $\theta_j$ in $G_{j-1}$. Next, a joint density is defined by the product of all removed factors as   
\begin{equation} \label{eq: joint density}
		\mathbb{P}_{joint} \left(\theta_j, S_j | D_j \right) \! = \eta_j^{-1} \! \prod_{f_i \in \mathcal{F}_{j-1}(\theta_j) } \!\!\!\! f_i (\Theta_i),
\end{equation}
where $\eta_j^{-1}$ is a normalizing term and $D_j$ represents all available data given by all factors removed up to and including the elimination of $\theta_j$, i.e. all data in $\mathcal{F} \setminus \mathcal{F}_{j}$. Using the chain rule, the joint density \eqref{eq: joint density} is then factorized as
\begin{equation} \label{eq:f_joint}
	\mathbb{P}_{joint} \left(\theta_j, S_j | D_j \right) = \prob{\theta_j | S_j, D_j} f_{new} \left(S_j | D_j \right).
\end{equation}
The conditional $\prob{\theta_j | S_j, D_j}$ is added as a new node to the Bayes net and the factor $f_{new} \left(S_j | D_j \right)$ is added into the factor graph $G_j$. Formally, each $G_j = \left( \mathcal{F}_j, \Theta_j, \mathcal{E}_j \right)$ is recursively defined by
\begin{align*}
	\mathcal{F}_j &= \mathcal{F}_{j-1} \setminus \mathcal{F}_{j-1}(\theta_j) \cup \{f_{new} \left(S_j | D_j \right)\}, \\
	\Theta_j &= \Theta_{j-1} \setminus \{ \theta_j \}, \\
	\mathcal{E}_j &= \mathcal{E}_{j-1} \setminus \mathcal{E}_{j-1}(\theta_j) \cup \{e(f_{new} \left(S_j | D_j \right), \theta) \; | \; \theta \in S_j \}.
\end{align*}
Note that in each intermediate step we have both an incomplete Bayes net and a reduced factor graph which defines a density on the remaining variables. 

Once all variables were eliminated the \forward pass is completed. The joint distribution \eqref{eq: posterior distribution} can then be expressed as the product of conditionals produced in each step as
\begin{equation} \label{eq: joint belief by conditionals}
	\prob{\Theta | D} = \prod_{j=1}^{|\Theta|} \prob{\theta_j | S_j, D_j},
\end{equation}
which defines the Bayes net (Fig. \ref{fig: bayes net}). The marginal distribution $\prob{\theta_j | D}$,
for each variable $\theta_j$, is calculated using backsubstitution in a second, \textit{backward} pass. Starting from the last eliminated variable and following the elimination order $\mathcal{O}$ in reverse, 
each marginal $\prob{\theta_j | D}$ is calculated by integrating over $S_j$ 
\small
\begin{equation} \label{eq: theoretical marginal theta_j}
	\footnotesize
	\prob{\theta_j | D} = \int_{S_j} \prob{\theta_j, S_j | D} \, dS_j  \nonumber \\
	= \int_{S_j} \prob{\theta_j | S_j , D} \cdot \prob{S_j | D} \, dS_j.
\end{equation}
\normalsize
Note that by definition, the separator for the last eliminated variable is an empty set, thereby yielding the direct marginal of the last eliminated variable. A key observation is that due to conditional independence $\prob{\theta_j | S_j , D} = \prob{\theta_j | S_j , D_j}$. As such, each marginal can be rewritten as
\begin{equation} \label{eq: theoretical marginal theta_j with conditional independance}
	\prob{\theta_j | D} = \int_{S_j} \prob{\theta_j | S_j , D_j} \cdot \prob{S_j | D} \, dS_j.
\end{equation}
In this work, we put forth a solution that extends to cases where no closed form solutions for calculating the marginal posterior distributions exist and the noise terms are non-Gaussian. In such cases, posterior distributions must be approximated. 

%% file: 04-methodology.tex
\section{METHODOLOGY}
Our key observation is that a joint probability function, for several random variables, can be seen as a \emph{high-dimensional surface} from which conditional and marginal distributions can be calculated. For example, consider the joint distribution $\prob{X,Y}$ over two random variables $X$ and $Y$. Using the chain rule, we rewrite the joint distribution as 
\begin{equation}
	\prob{X,Y} = \prob{X | Y} \cdot \prob{Y},
\end{equation}
where the conditional $\prob{X | Y}$ is given by a specific \slice from the high-dimensional surface for each realization of $Y$. See illustration in Fig \ref{fig: surface with slices}.

The marginal distribution $\prob{X}$ is calculated using the Chapman-Kolmogorov transit integral 
\begin{equation} \label{eq: generic marginal}
	\prob{X} = \int_{Y} \prob{X | Y} \cdot \prob{Y} \, dY. 
\end{equation}
As there is no closed form solution for \eqref{eq: generic marginal} in the general case, it can be approximated using $N$ samples of $Y$ from $\mathbb{P}(Y)$. The corresponding estimated marginal is given by 

\begin{equation} \label{eq: approx marginal}
	\hat{\mathbb{P}} \left( X \right) 
	= \!\! \underset{y\sim \prob{Y}}{\hat{\mathbb{E}}}[\prob{X | Y=y}] \! = \!
	\frac{1}{N} \sum_{i=1}^{N} \prob{X | Y=y^i}.
\end{equation} 
Put differently, evaluating \eqref{eq: approx marginal} can be seen as accessing the high-dimensional surface at specific sample points of $Y$ in order to extract a \mixture of conditional \slices over $X$. Moreover, when $N\rightarrow \infty$ this approximation is guaranteed to converge to the true analytical solution. 
 

\subsection{Inference Using Slices} \label{sec: inference using slices}
For nonparametric inference, our proposed approach leverages the \fb algorithm presented in Sec. \ref{sec: background}. For each eliminated variable $\theta_j$ in the \forward pass, the joint density defined in \eqref{eq: joint density} is factorized into a conditional and a marginal via \eqref{eq:f_joint} which is rewritten once again for convenience
\begin{equation} \label{eq:f_joint2}
	\mathbb{P} \left(\theta_j, S_j| D_j \right) = \prob{\theta_j | S_j, D_j} f_{new} \left(S_j | D_j \right).
\end{equation}
This factorization serves as the pivotal step through which information propagates across the algorithm. In strike contrast to existing approaches that depend on intermediate density reconstructions, our primary contribution is the direct use of \slices to approximate both $f_{new} \left(S_j | D_j \right)$ and the conditional $\prob{\theta_j | S_j, D_j}$. We rigorously describe how our \slices perspective is systematically employed throughout the algorithm.

\subsubsection{Forward Pass} \label{sec: forward pass}
We start by explicitly writing  
the term $f_{new} \left(S_j | D_j \right) = \int_{\theta_j} \mathbb{P}_{joint} \left(\theta_j, S_j | D_j \right) d\theta_j$ in \eqref{eq:f_joint2} as
\begin{equation} \label{eq: f_new integral}
	f_{new} \left(S_j | D_j \right) = \eta_j^{-1}\int_{\theta_j} \prod_{f_i \in \mathcal{F}_{j-1}(\theta_j)} \!\!\!\! f_i (\Theta_i) \;\; d\theta_j,
\end{equation}
where $\eta_j^{-1}$ is a normalizing term. As there is no closed form solution for \eqref{eq: f_new integral}, it must be approximated. While other approaches use additional intermediate approximations,
such as KDE in the case of \cite{Fourie16iros} or learning techniques as seen in \cite{Huang23tro}, we employ \slices as a more direct and effective means of approximation. We note that we assume the ability to access such high-dimensional \slices, an assumption shared with mm-iSAM, NF-iSAM, and NSFG. By generating $N$ samples of  $\theta_j$ from some factor $f' \in  \mathcal{F}_{j-1} (\theta_j)$, a standard Monte Carlo estimator of \eqref{eq: f_new integral} is given by
\begin{equation} \label{eq: f_new approx}
	\hat{f}_{new} \left(S_j | D_j \right)  =   \frac{\eta_j^{-1}}{N} \! \!\sum_{n=1}^{N} \! \prod_{f_i \in \mathcal{F}_{j-1}(\theta_j) \setminus \{f'\}} \!\! \!\! \!\!\!\!\!\! \!\!\!\!f_i \left(\theta_j^n, \Theta_i \! \setminus \! \{\theta_j \} \right),
\end{equation}
where $\Theta_i \setminus \{\theta_j \} \subseteq S_j $ and each $f_i \in \mathcal{F}_{j-1}(\theta_j) \setminus \{f' \}$ is evaluated using the samples of $\theta_j$.
In a \slices perspective, for each sample $\theta_j^n$, a high-dimensional conditional \slice over $\Theta_i \setminus \{\theta_j\}$ is retrieved from every non-unary factor $f_i \in \mathcal{F}_{j-1}(\theta_j)$. We acknowledge a slight misuse of notation, as $D_j$ is no longer purely theoretical due to the fact that the conditioned data is approximated at each step by $\hat{f}_{new}$ (see \textbf{Example} below). 

\begin{figure} [h]
	\centering
	\begin{subfigure}[b]{0.25\textwidth}
			\centering
			\includegraphics[scale=0.32]{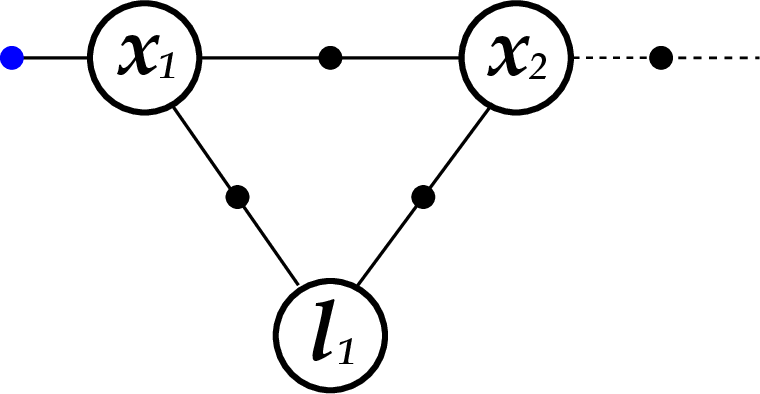}
			\caption{}
			\label{fig: elimination example a}
	\end{subfigure}
	\begin{subfigure}[b]{0.22\textwidth}
			\centering
			\includegraphics[scale=0.32]{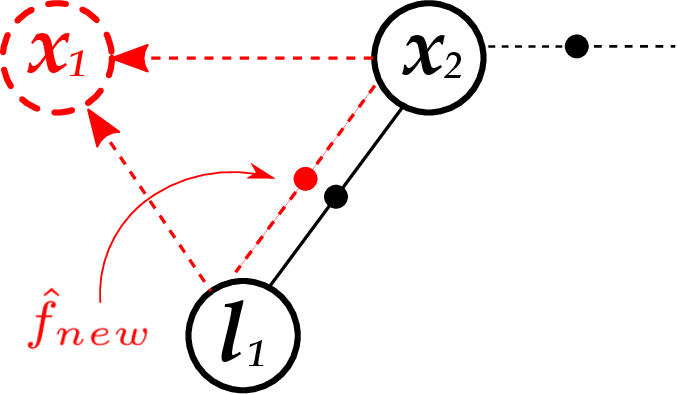}
			\caption{}
			\label{fig: no unaray l_1}
	\end{subfigure}
	\caption{\scriptsize Generating samples during variable elimination; (a) A subset of three variables $x_1,x_2,l_1$ from a larger factor graph. The given elimination order is $\mathcal{O} = \{\theta_1=x_1, \theta_2=l_1, ...\}$. When eliminating $x_1$, samples of $x_1$ are directly obtained from \textcolor{blue}{$f(x_1)$} to approximate the new factor \textcolor{red}{$\hat{f}_{new} (l_1, x_2|D_1)$} with \slices via (13);   
	(b) When eliminating $l_1$, samples of $l_1$ are obtained according to Lemma \ref{lemma: non unary factors} from \textcolor{red}{$\hat{f}_{new} (l_1, x_2|D_1)$}. Even without a unary factor on $l_1$, our \slices approach provides a method to directly generate samples of $l_1$.}
	\label{fig: no unary}
\end{figure}

Obtaining samples of $\theta_j$ is essential
for approximating \eqref{eq: f_new integral}. However, generating these samples becomes non-trivial when no unary factor $f\in \mathcal{F}_{j-1}(\theta_j)$ is available for direct sampling. To handle this issue, mm-iSAM employs multiscale Gibbs sampling, which is computationally expensive. In contrast, both NF-iSAM and NSFG break the factor graph to chain like structures and generate samples using ancestral sampling, which does not leverage loop closure factors. Through direct approximation of \eqref{eq: f_new integral} with \slices in each step, our approach eliminates the need for iterative procedures or factor graph decomposition into chain-like structures for generating samples of $\theta_n$. Unlike previous methods, this holds true even in the absence of unary factors. 

\begin{lemma} \label{lemma: non unary factors}
	Given a factor graph $G = \left( \mathcal{F}, \Theta, \mathcal{E} \right)$ and an elimination order $\mathcal{O}$, if each eliminated variable $\theta_j \in \Theta$ either has a unary factor connected to it, i.e. $\exists f(\theta_j) \in \mathcal{F}_{j-1} (\theta_j)$ or, $\exists \theta_i \in \Theta$ such that $\theta_i$ was previously eliminated and $f(\theta_i, \theta_j) \in \mathcal{F}$, then samples of $\theta_j$ can be drawn from one of the factors $\mathcal{F}_{j-1} (\theta_j)$ (see proof in appendix).
\end{lemma} 

\textbf{Example: }\emph{We demonstrate how Lemma \ref{lemma: non unary factors} is utilized to draw samples when there are no unary factors using the example in Fig. \ref{fig: no unary}. 
After eliminating the first variable $\theta_1 = x_1$, a new factor $\hat{f}_{new} (l_1, x_2 | D_1)$ is added to the factor graph.  Consequently, the second variable to be eliminated, $\theta_2=l_1$, lacks any unary factors at this stage. According to \eqref{eq: joint density}, the joint density $\mathbb{P}_{joint}(l_1,x_2|D_2)$ is given by
\begin{equation} \label{eq: example joint}
	\mathbb{P}_{joint} \left(l_1, x_2| D_2 \right) = \hat{f}_{new} \left(l_1, x_2 | D_1 \right) \cdot f(l_1, x_2),
\end{equation}
and is factorized, according to \eqref{eq:f_joint}, as
\begin{equation}
	\mathbb{P}_{joint} \left(l_1, x_2| D_2 \right) = \prob{l_1 | x_2, D_2} \cdot f_{new} \left(x_2 | D_2 \right),
\end{equation}
given the separator $S_2=\{x_2\}$. As previously stated, it is important to note that $D_2$ does not encompass all raw data up to this point. Instead of $D_2=\{b_0, u_1, z_1, z_2\}$, where $b_0\equiv f(x_1)$ is given by the prior, we have $D_2=\{ \hat{f}_{new} \left(l_1, x_2 | D_1 \right), z_2 \}$ due to the approximation from the previous elimination step.
We write $f_{new} \left(x_2 | D_2 \right)$ explicitly, following \eqref{eq: f_new integral} and considering the estimator $\hat{f}_{new} (l_1, x_2 | D_1)$ obtained via \eqref{eq: f_new approx}. Denoting it intermediately by $\tilde{f}_{new} (x_2|D_2)$ and changing the order of the integral and summation, yields 
\begin{equation} \label{eq: f_new(x_2) explicit}
	\small
	\tilde{f}_{new} \left(x_2 | D_2 \right) = \frac{\eta^{-1}}{N} \sum_{n=1}^{N} f \left(x_1^n, x_2 \right) \!\! \int_{l_1} \!\! f \left(x_1^n, l_1 \right) \cdot f \left(l_1, x_2 \right) \; dl_1.
\end{equation}
We can now exploit the unary structure of $f \left(x_1^n, l_1 \right)$, for each $x_1^n$, to generate samples of $l_1$ and approximate \eqref{eq: f_new(x_2) explicit} as 
\begin{equation*}
	\hat{f}_{new} \left(x_2 | D_2 \right) =
	 \frac{\eta^{-1}}{N^2} \sum_{n_1=1}^{N} f \left(x_1^{n_1}, x_2 \right) \!\! \sum_{n_2=1}^{N} f \left(l_1^{n_1,n_2}, x_2 \right).
\end{equation*}
Unlike other methods, as previously discussed, our \slices approach exclusively utilizes structure for sample generation when there are no unary factors.}

\vspace{5pt}
We next turn our attention to the conditional term $\prob{\theta_j | S_j, D_j}$ in \eqref{eq:f_joint2}. By utilizing $\hat{f}_{new}$, given by \eqref{eq: f_new approx}, and replacing  \eqref{eq: joint density}, an estimator of the conditional is obtained
\begin{equation}
	\small \label{eq:prop_conditional_est_specific}
	\hat{\mathbb{P}}(\theta_j | S_j, D_j)  = \frac{\prod_{f_i \in \mathcal{F}_{j-1}(\theta_j)} f_i \left(\Theta_i \right)}{\frac{1}{N} \sum_{n=1}^{N} \prod_{f_i \in \mathcal{F}_{j-1}(\theta_j) \setminus \{f'\}} f_i \left(\theta_j^n, \Theta_i \setminus \{ \theta_j \} \right)}.
\end{equation}
We note that the normalization term $\eta_j^{-1}$ cancels out. Moreover, 
there is no need to explicitly calculate $\eta_j^{-1}$ in \eqref{eq: f_new approx} as intermediate new factors created during the \forward pass are only used to generate samples of $S_j$. During the \backward pass, \slices from the high-dimensional surface representing the numerator, are used to approximate the conditional \eqref{eq:prop_conditional_est_specific} as described next.

\subsubsection{Backward Pass}
Once the \forward pass concludes, we utilize a similar \slices perspective, leveraging the constructed Bayes net, to retrieve the joint and marginal posterior distributions. Following the elimination order $\mathcal{O}$ in reverse, we obtain an 
estimator of the marginal \eqref{eq: theoretical marginal theta_j}
for each variable $\theta_j$. This is achieved by utilizing the estimator of the conditional $\prob{\theta_j | S_j , D_j}$, which was already evaluated during the \forward pass 
\eqref{eq:prop_conditional_est_specific}, and samples of $S_j$ directly obtained from the marginal $\hat{\mathbb{P}}(S_j | D)$ which is given by a \mixture of \slices.
Specifically, given $N$ samples of $S_j \sim \hat{\mathbb{P}}(S_j | D)$, the estimator is given by
\begin{equation} \label{eq: approximated marginal}
	\hat{\mathbb{P}}(\theta_j | D)  = \frac{1}{N} \sum_{n=1}^{N} \hat{\mathbb{P}}(\theta_j | S_j^n, D_j),
\end{equation}
which is by itself also a \mixture of high-dimensional \slices. For each sample $S_j^n$, a high-dimensional conditional \slice over $\theta_j$ is retrieved from every non-unary factor $f_i \in \mathcal{F}_{j-1}(\theta_j)$. The weight of each \slice is given by evaluating the denominator of \eqref{eq:prop_conditional_est_specific} at sample point $S_j^n$. We note that \eqref{eq: approximated marginal} represents the marginal as a distribution. As such, we can both sample from the marginal and evaluate the likelihood for a given sample.

Similarly, the joint posterior distribution \eqref{eq: belief} can also be approximated. By replacing each conditional in \eqref{eq: joint belief by conditionals} with \eqref{eq:prop_conditional_est_specific}, we can explicitly represent the joint distribution from which we can also both sample and evaluate the likelihood for a given sample. Sampling from the joint distribution is done by following the elimination order $\mathcal{O}$ in reverse. We first generate a sample of the last eliminated variable directly from the corresponding marginal. Next, we recursively sample each variable from the corresponding conditional \eqref{eq:prop_conditional_est_specific}, given a sample of the separator, to obtain a single sample from the joint distribution. 

Up to this point, our approach has focused on a batch processing method applied to a given factor graph. However, in real-world applications, a real-time online solution is needed. This requires continuous updates and providing estimates whenever new measurements are added. 

\subsection{Incremental Inference with Early Stopping Heuristic}
Incremental inference was initially developed for the Gaussian case \cite{Dellaert06ijrr}, \cite{Kaess08tro}, \cite{Kaess12ijrr}. The underlining observation in these approaches is that the addition of new measurements affects only specific parts the original factor graphs.
As a result, there's no need to redo the entire \fb inference process. Instead, an incremental \forward pass is conducted, focusing computations solely on the affected variables. This process, termed re-elimination, significantly reduces computational costs by reusing previously computed information and selectively updating relevant sections of the factor graph (See Algorithm 1 in \cite{Indelman15ras} for a simplified version). Furthermore, performing full backsubstitution in every iteration can be computationally expensive, particularly in large-scale problems with complex factor graphs. Consequently, iSAM2 \cite{Kaess12ijrr} incorporated an early stopping heuristic in the downward pass to significantly reduce computational cost in the non-linear Gaussian case.

By utilizing the Bayes tree \cite{Kaess12ijrr}, both mm-iSAM \cite{Fourie16iros} and NF-iSAM \cite{Huang23tro} achieve nonparametric incremental inference capabilities. Nevertheless, these algorithms restrict the application of incremental aspects solely to the upward pass, while the entire downward pass is executed anew. Moreover, training new conditional samplers for affected variables in NF-iSAM, during re-elimination, must be performed from scratch, resulting in computational inefficiency.

To integrate incremental aspects into both the \forward and \backward passes, we suggest the adoption of re-elimination and an early stopping heuristic. This combination, which serves to notably diminish computational complexity, is a novel concept introduced for the first time in a nonparametric setting to the best of our knowledge.

We employ the Maximum Mean Discrepancy (MMD) \cite{Gretton12jmlr} metric as a heuristic for early stopping of the \backward pass. The MMD is a metric used to assess the dissimilarity between two probability distributions, relying on samples. During the \backward pass of each incremental step, we calculate and cache the most recent marginal distributions. Given a new marginal and a previously found marginal for a given variable, we use the MMD metric to evaluate the distance between the two distributions. If this distance is below a user defined threshold, we stop the \backward pass. Notably, we have two hyper-parameters available for fine-tuning, providing a balance between accuracy and efficiency: the number of samples used for evaluating the MMD distance and the threshold criteria.

%% file: 05-results.tex
\section{Experiments}
We evaluate our proposed \slices approach using both synthetic and real-world datasets. To assess its performance, we compare our results with those obtained using mm-iSAM \cite{Fourie16iros}, NF-iSAM \cite{Huang23tro}, and NSFG \cite{Huang22ral}, employing the open-source code provided by the respective authors. Following \cite{Huang22ral} and \cite{Huang23tro}, we use the root mean square error (RMSE) metric to measure the discrepancies between ground truth and samples from posterior distributions. For robustness, all reported results for each method were averaged over ten independent runs.

\subsection{Synthetic Dataset - Multi Modal Four Doors}
The synthetic four doors localization example \cite{Fourie16iros} is a one-dimensional SLAM problem. The robot is aware of a map with four identical doors but initially observes only one of them, without knowing which specific door it is. Consequently, the prior distribution in this scenario is multi-modal, with four modes of equal weight. As the robot progresses through several steps, it gathers loop closure measurements related to a landmark within its environment. Additionally, it acquires two more measurements, each corresponding to different doors. These latter measurements play a critical role in resolving the ambiguity, leading to the collapse of the posterior distribution into a single mode.

\begin{figure} [h]
	\centering
	\includegraphics[scale=0.29]{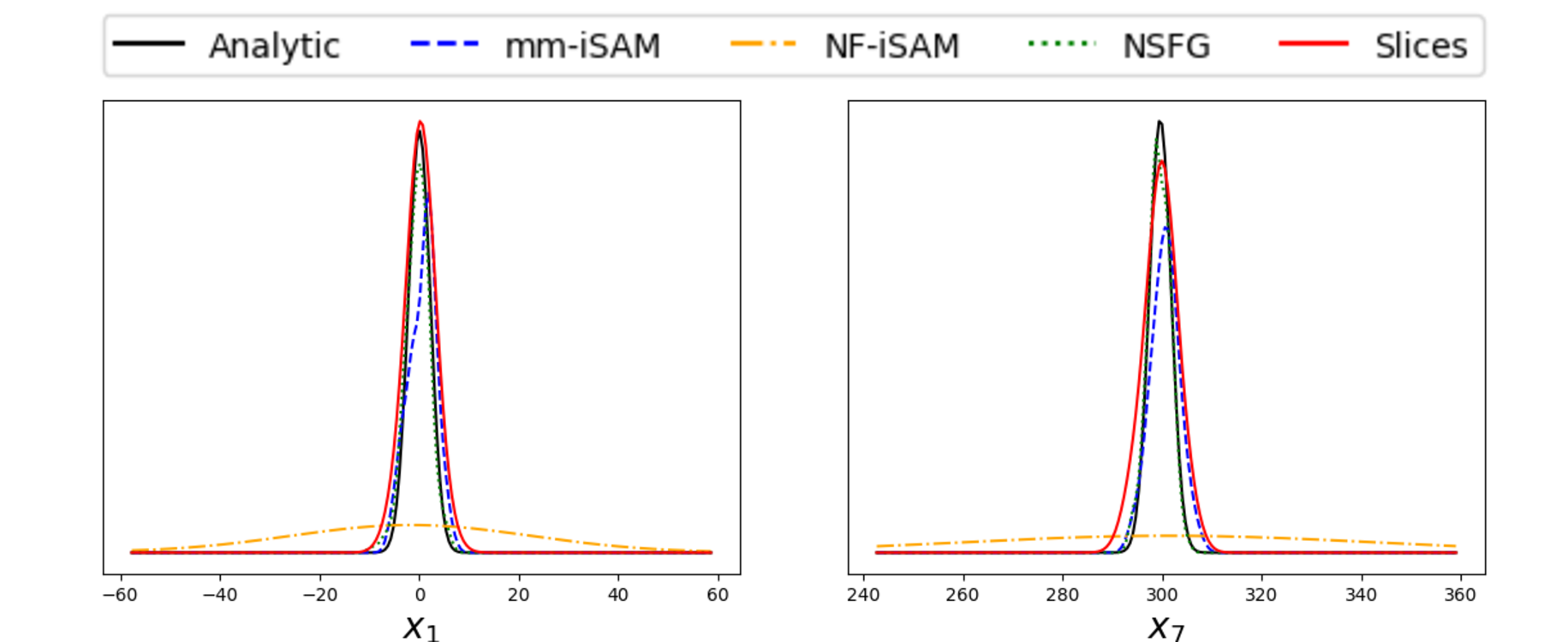}
	\caption{Four doors synthetic SLAM dataset \cite{Fourie16iros} marginal posterior distributions for the first and last robot poses. A black solid line indicates the analytic solution when considering only the mode which corresponds to the ground
	truth. Our \slices approach directly approximates these marginal via (17). 
	In mm-ISAM, KDEs are used to approximate these marginal distributions, whereas in NF-iSAM and NSFG, they are derived solely from samples and approximated using KDEs based on these samples, as demonstrated in \cite{Huang22ral}. In each method, 200 samples were utilized to approximate the distributions.}
	\label{fig: four doors marginals}
\end{figure}

\begin{figure} [h]
	\centering
	\begin{subfigure}[b]{0.50\columnwidth}
		\includegraphics[scale=0.28]{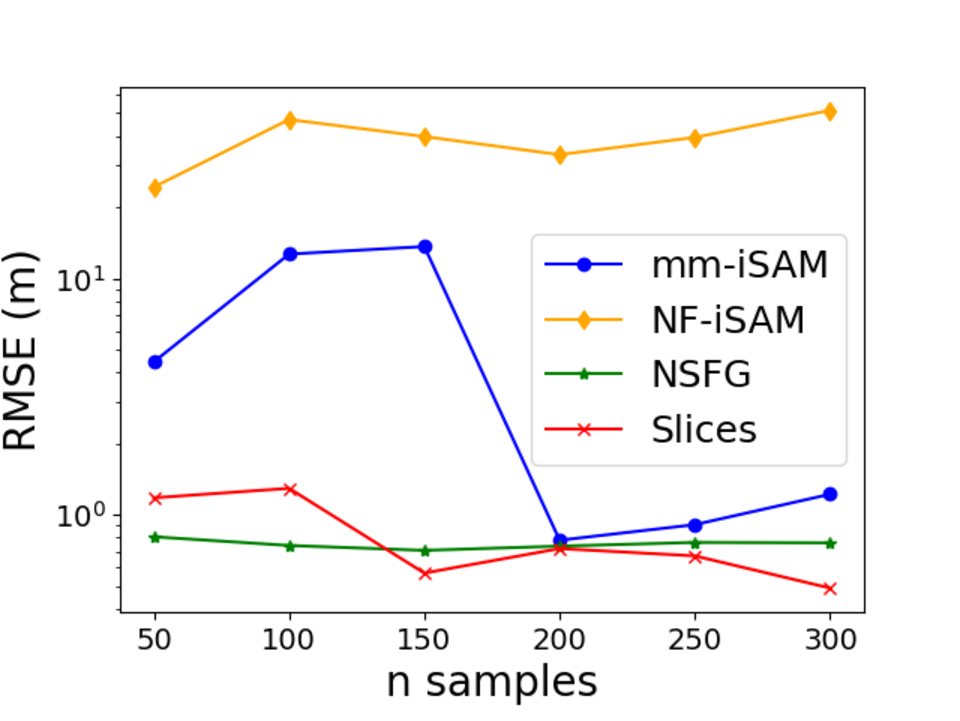}
		\caption{}
		\label{fig: 4_doors_landmark_rmse}
	\end{subfigure}
	\begin{subfigure}[b]{0.48\columnwidth}
		\includegraphics[scale=0.28]{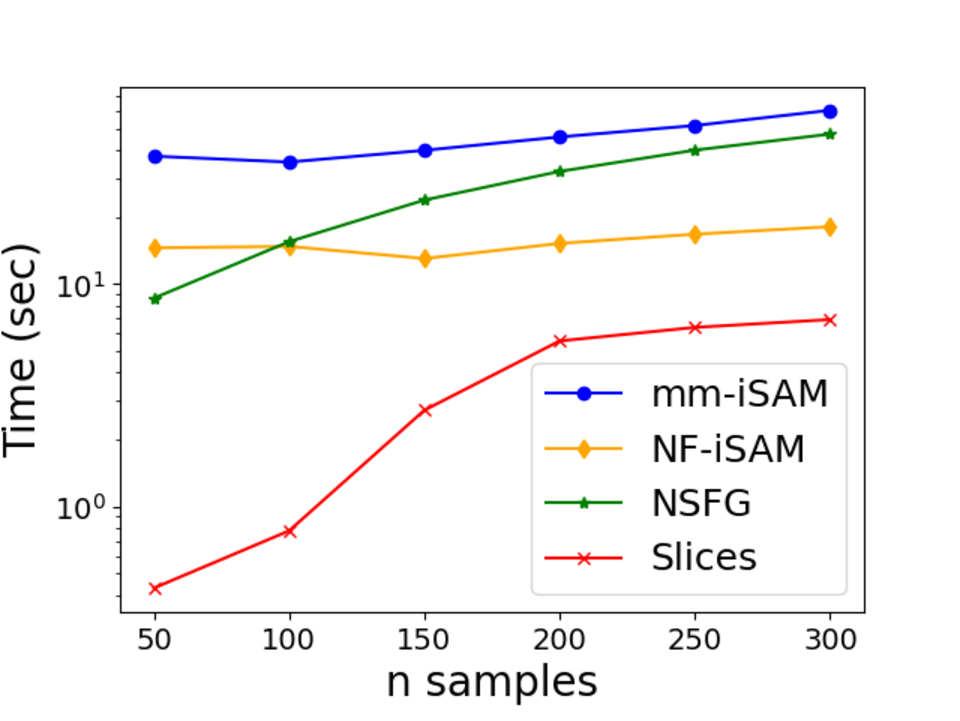}
		\caption{}
		\label{fig: 4_doors_landmark_time}
	\end{subfigure}
	\caption{Four doors synthetic SLAM dataset \cite{Fourie16iros} results. We report RMSE (m) and run time (sec) as functions of the number of samples used by all methods. \scriptsize}
	\label{fig: four doors}
\end{figure}

This one dimensional problem is interesting as the marginal posterior distributions for all robot poses and the landmark can be analytically calculated (as discussed in \cite{Fourie16iros}). 
Our \slices approach directly approximates
these marginal distributions via \eqref{eq: approximated marginal}. In mm-ISAM these marginal distributions are approximated with KDEs while in NS-iSAM and NSFG they are only available through samples (Fig. \ref{fig: four doors marginals}). Note how NF-ISAM fails to recover the correct mode, even in this small scenario, due to its difficulty in converging when the number of samples is relatively small. 

In this experiment, we assess run time and RMSE as functions of the number of samples generated by the algorithms in each step, employing a batch computation approach across the entire factor graph. Our \slices approach demonstrates superior accuracy compared to both mm-iSAM and NF-iSAM, achieving results on par with NSFG (Fig. \ref{fig: four doors}), despite the latter being recognized as an offline algorithm. Furthermore, our approach shows significantly improved computational efficiency when using the same number of samples.

\subsection{Real World Dataset - Plaza}
The Plaza2 dataset \cite{Djugash09jfr} comprises a sequence of odometry and range measurements to four unknown landmarks gathered by a vehicle navigating a planar environment. 
This range measurement SLAM problem is challenging because it involves both nonlinear measurements and non-Gaussian likelihood models. 

In this experiment, we perform incremental inference and evaluate both the run time and RMSE at each step.
We adopt the same noise models across all methods, following the specifications outlined in \cite{Huang23tro}, for a fair compression. We also set our hyperparameters, for the early stopping heuristic, to $N_M=100$ and $\delta=1e^{-4}$, representing the number of samples used for evaluating the MMD distance and the threshold criteria, respectively. The chosen re-elimination order for affected variables, in each incremental step, is such that landmark variables are eliminated last. 

\begin{figure} [h]
	\centering
	\includegraphics[scale=0.19]{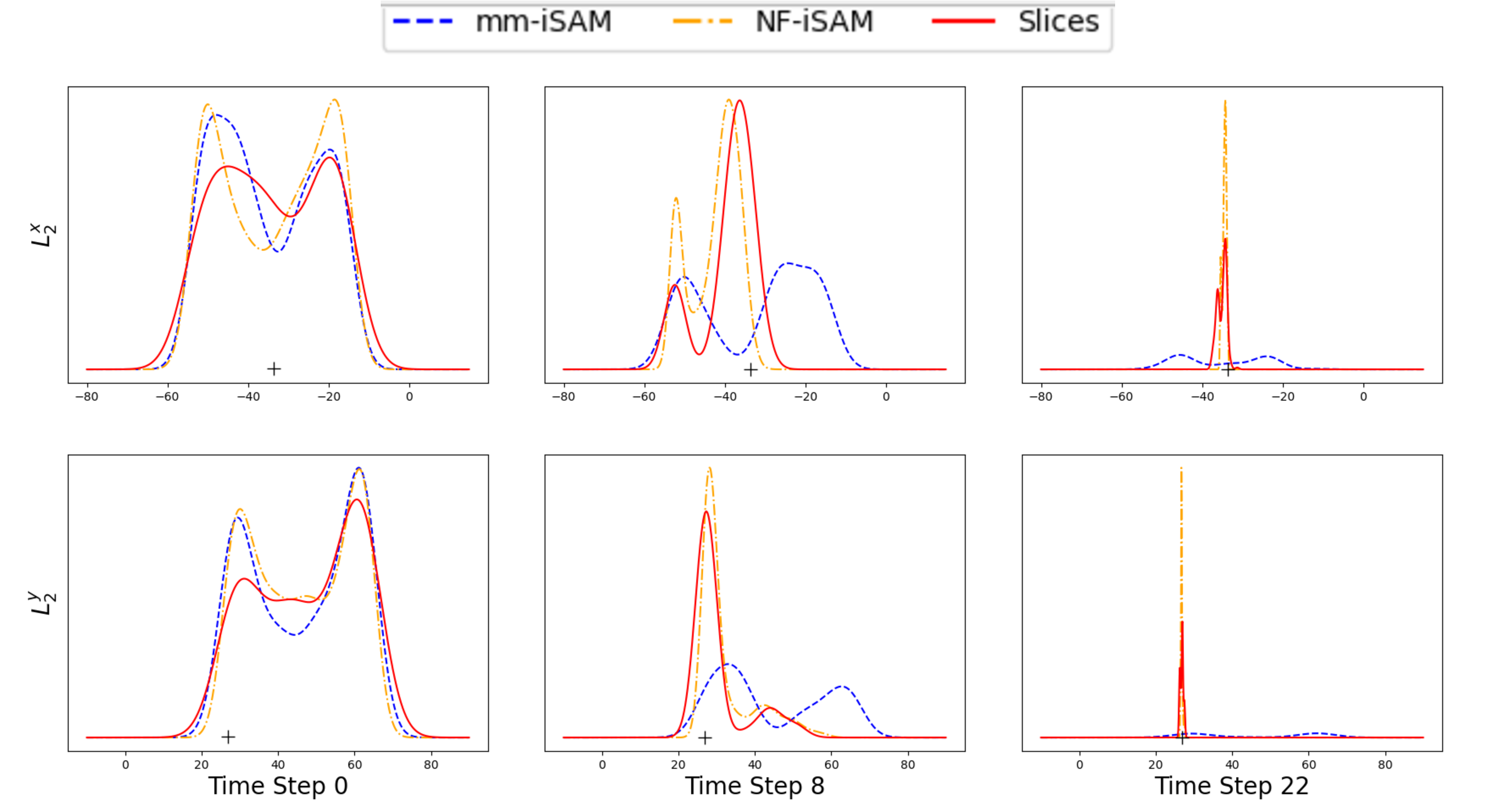}
	\caption{Plaza2 dataset \cite{Djugash09jfr} marginal posterior distributions for landmark number 2 across different time steps, presented separately for the $x$ and $y$ axes. The robot captures the first, fifth and ninth range measurement to $L_2$ at time steps 0, 8 and 22 respectively. The ground truth position of $L_2$ is denoted by a black $+$ sign. The hyperparameters used for each method are described in Section V.B .}
	\label{fig: plaza2 L2 marginals}
\end{figure}

\begin{figure} [h]
	\centering
	\begin{subfigure}[b]{0.50\columnwidth}
		\includegraphics[scale=0.28]{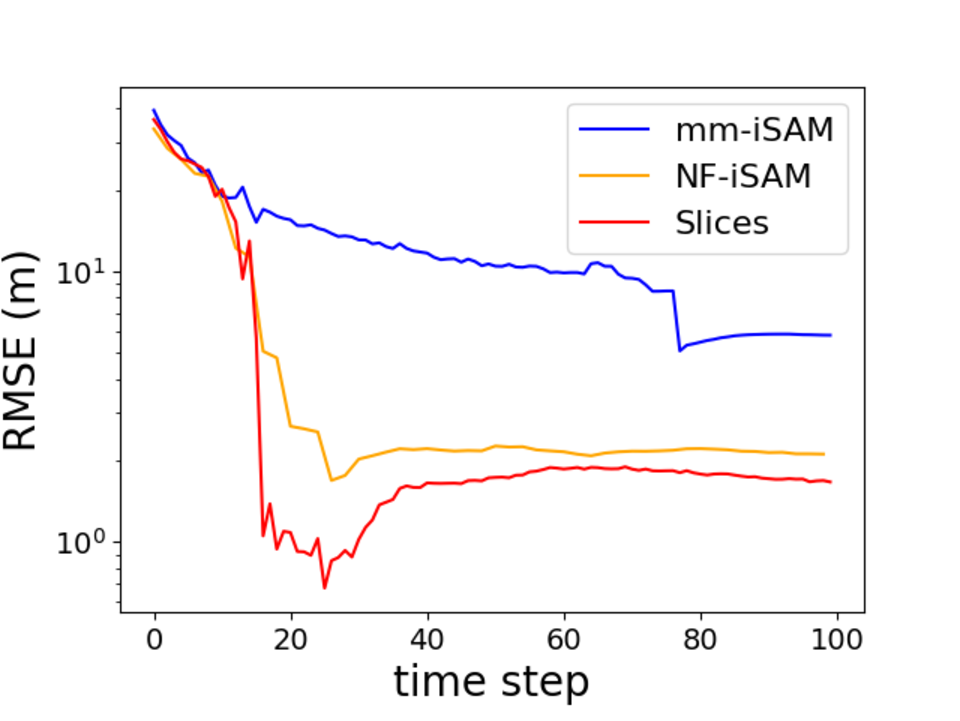}
		\caption{}
		\label{fig: plaza2_rmse}
	\end{subfigure}
	\begin{subfigure}[b]{0.48\columnwidth}
		\includegraphics[scale=0.28]{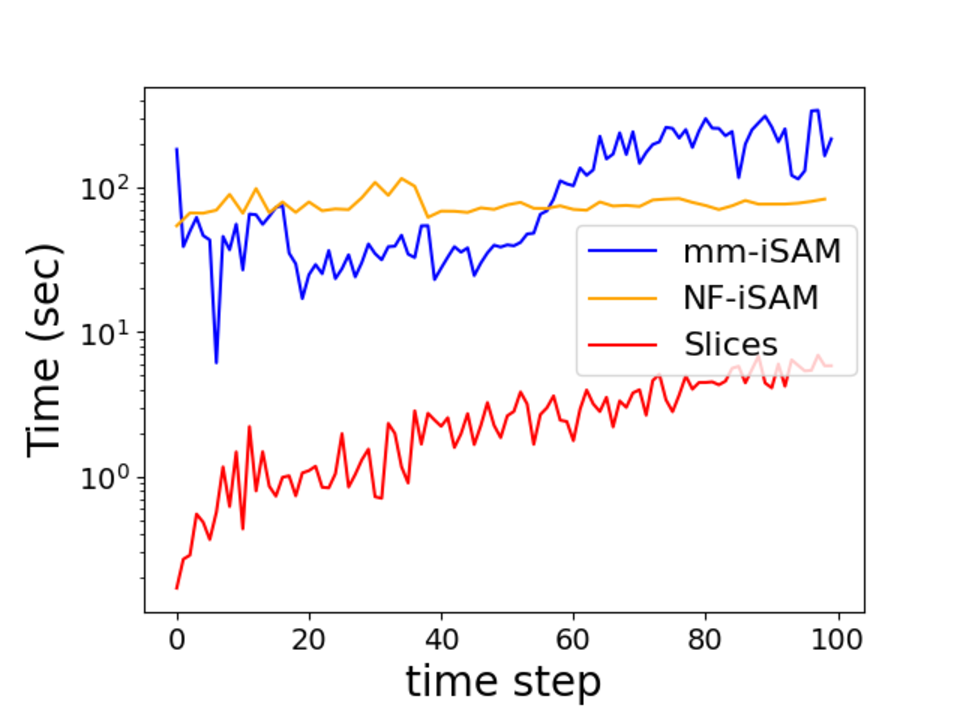}
		\caption{}
		\label{fig: plaza2_time}
	\end{subfigure}
	\caption{Plaza2 dataset \cite{Djugash09jfr} with range measurements. We report RMSE (m) and run time (sec) for each incremental step. \scriptsize}
	\label{fig: plaza2}
\end{figure}

We configured our \slices approach to generate 150 samples in each step. For NF-iSAM, we followed the recommended setting of 2000 samples and used the same hyperparameters as published by the authors in \cite{Huang23tro}. As for mm-iSAM, we employed the default value of $100$ samples, as advised in the open source implementation of \cite{Fourie16iros}. Notably, when we attempted to use 200 samples, the mm-iSAM run time became prohibitively long for this specific problem. The scale and dimensionality of this problem were too demanding for NSFG, which is also confirmed in \cite{Huang22ral},   where the authors only presented results for a limited number of time steps. Notably, our \slices approach achieves superior accuracy compared to NF-iSAM and mm-iSAM (Fig. \ref{fig: plaza2}). Furthermore, our method significantly enhances computational efficiency, reducing complexity by an order of magnitude compared to NF-iSAM and mm-iSAM. An illustration depicting the marginal distribution across different time steps for the second landmark is shown in Fig. \ref{fig: plaza2 L2 marginals}.

%% file: 06-conclusions.tex
\section{CONCLUSIONS}
In this work, we introduced a novel approach that utilizes \slices from high-dimensional surfaces to efficiently approximate nonparametric posterior distributions. Unlike existing methods that use generated samples to reconstruct intermediate distributions through learning procedures and KDEs, our \slices perspective avoids such additional intermediate approximations. Moreover, our approach does not use any iterative procedures as in \cite{Fourie16iros, Huang22ral, Huang23tro}.

We have also introduced a novel early stopping heuristic, a first in nonparametric scenarios, during the \backward pass to significantly reduce computational complexity and enable real-time operation. In future research, we plan to investigate the impacts of different thresholds for the early stopping heuristic.

Our \slices perspective not only demonstrated superior accuracy compared to other online nonparametric inference methods \cite{Fourie16iros, Huang23tro}, but also matched the performance of an offline, state-of-the-art, nonparametric inference method \cite{Huang22ral}. Additionally, our approach achieved a remarkable reduction in computational complexity, as evidenced in both synthetic and real-world datasets.

%% file: 07-appendix.tex
\section{APPENDIX: Proof of Lemma \ref{lemma: non unary factors}}
\label{app:marginal}
We prove Lemma \ref{lemma: non unary factors}  by induction. \\
\textit{\underline{base case:}} Let $\theta_1$ be the first eliminated variable. Due to the properties of the elimination order given in Sec.  \ref{sec: background}, $\exists f(\theta_1) \in \mathcal{F}$ from which samples of $\theta_1$ can be generated. \\
\textit{\underline{induction step:}} Eliminating $\theta_j$, if $\exists f(\theta_j) \in \mathcal{F}$ then according to the elimination algorithm it holds that $f(\theta_j) \in \mathcal{F}_{j-1}$ from which samples of $\theta_j$ can be generated. Else, according to the given elimination order $\exists \theta_i \in \Theta$ such that $\theta_i$ was previously eliminated and $f(\theta_i, \theta_j) \in \mathcal{F}$. Since $f(\theta_i, \theta_j) \in \mathcal{F}$, according to the elimination algorithm it holds that $f(\theta_i, \theta_j) \in \mathcal{F}_{i-1}$ and thus $\theta_j \in S_i$. As such, according to \eqref{eq: f_new integral}, when $\theta_i$ was eliminated, a new factor 
\begin{equation} \label{eq: f_new theta_i}
	f_{new} \left(S_i | D_i \right) =
	\eta^{-1}\int_{\theta_i} \prod_{f_k \in \mathcal{F}_{i-1}(\theta_i)} f_k \left(\Theta_k \right) d\theta_i,
\end{equation}
was added. Using the induction assumption, samples of $\theta_i$ were generated from one of the factors in $\mathcal{F}_{i-1}(\theta_i)$ to approximate this integral. 
Without loss of generality, we denote the factor from which samples of $\theta_i$ were generated as $f_{\bar{k}}$ and write the approximation to \eqref{eq: f_new theta_i} as
\begin{equation} \small
	\hat{f}_{new} \left(S_i | D_i \right)  =   \frac{\eta^{-1}}{N} \sum_{n=1}^{N} f(\theta_i^n, \theta_j) \!\! \!\! \!\ \!\! \!\!\!\!\!\!\!\!\!\!\!\!\!\ \prod_{f_k \in \mathcal{F}_{i-1}(\theta_i) \setminus \{f(\theta_i^n, \theta_j), f_{\bar{k}}\}}  \!\! \!\! \!\!\!\!\!\! \!\!\!\!\!\!\! \!\!\!\!\!\!\! \! f_k \left(\theta_i^n, \Theta_k^{\neg i} \right).
\end{equation}
According to the elimination algorithm, for any variable $\theta_m \in S_i$ eliminated after $\theta_i$ and before $\theta_j$, it must hold that the new factor $f_{new} \left(S_m | D_m \right)$ must contain $f(\theta_i^n, \theta_j)$. Thus, $ \mathcal{F}_{j-1}$ also contains $f(\theta_i^n, \theta_j)$ from which samples of $\theta_j$ can be generated. $\blacksquare$